\documentclass[conference]{IEEEtran}
\makeatletter
\let\NAT@parse\undefined
\makeatother

\IEEEoverridecommandlockouts                              

\usepackage{cite}
\usepackage{amsmath,amssymb,amsfonts}
\usepackage{algorithmic}
\usepackage{graphicx}
\usepackage{textcomp}
\usepackage{xcolor}
\usepackage{array}
\usepackage{multirow}
\usepackage{tabularx}
\usepackage{makecell}
\usepackage{subfig}
\usepackage[square,sort,comma,numbers]{natbib}
\usepackage[utf8]{inputenc}
\usepackage{lipsum}
\usepackage{eso-pic}

\AddToShipoutPictureFG*{%
  \ifnum\value{page}=1
    \put(50,40){%
      \parbox{\textwidth}{%
        \centering \footnotesize
        © 2025 IEEE. Personal use of this material is permitted.  Permission from IEEE must be obtained for all other uses, in any current or future media, including reprinting/republishing this material for advertising or promotional purposes, creating new collective works, for resale or redistribution to servers or lists, or reuse of any copyrighted component of this work in other works.
      }
    }
  \fi
}

\begin{document}

\title{\LARGE \bf
Engagement and Disclosures in LLM-Powered Cognitive Behavioral Therapy Exercises: A Factorial Design Comparing the Influence of \\a Robot vs. Chatbot Over Time
}

\author{Mina Kian$^{1}$, Mingyu Zong$^{1}$, Katrin Fischer$^{1}$, Anna-Maria Velentza$^{2}$, Abhyuday Singh$^{1}$, Kaleen Shrestha$^{1}$, \and Pau Sang$^{1}$, Shriya Upadhyay$^{1}$, Wallace Browning$^{1}$, Misha Arif Faruki$^{1}$, Sébastien M. R. Arnold$^{1}$, \and \makebox[\linewidth]{Bhaskar Krishnamachari$^{1}$, Maja Matarić$^{1}$}
\thanks{$^{1}$ Mina Kian, Mingyu Zong, Katrin Fischer, Abhyuday Singh, Kaleen Shrestha, Pau Sang, Shriya Upadhyay, Wallace Browning, Misha Arif Faruki, Sébastien M. R. Arnold, Bhaskar Krishnamachari, and Maja Matarić are with the University of Southern California, Los Angeles, California, USA.
        {\tt\small kian, mzong, katrinfi, asingh17, kshresth, psang, shriyaup, wbrownin, mfaruki, arnolds, bkrishna, mataric@usc.edu}}%
\thanks{$^{2}$Anna-Maria Velentza is with the Lab-STICC CNRS UMR 6285, Brest National Engineering School (ENIB), Brest, France.
         {\tt\small b.d.researcher@ieee.org}}%
}

\maketitle




\begin{abstract}
Many researchers are working to address the worldwide mental health crisis by developing therapeutic technologies that increase the accessibility of care, including leveraging large language model (LLM) capabilities in chatbots and socially assistive robots (SARs) used for therapeutic applications. Yet, the effects of these technologies over time remain unexplored. In this study, we use a factorial design to assess the impact of embodiment and time spent engaging in therapeutic exercises on participant disclosures. We assessed transcripts gathered from a two-week study in which 26 university student participants completed daily interactive Cognitive Behavioral Therapy (CBT) exercises in their residences using either an LLM-powered SAR or a disembodied chatbot. We evaluated the levels of active engagement and high intimacy of their disclosures (opinions, judgments, and emotions) during each session and over time. Our findings show significant interactions between time and embodiment for both outcome measures: participant engagement and intimacy increased over time in the physical robot condition, while both measures decreased in the chatbot condition.

\end{abstract}


\section{Introduction}
In the ongoing worldwide mental health crisis, university students are particularly vulnerable; the proportion of students exhibiting symptoms of one or more mental health conditions increased by nearly 50\% from 2013 to 2021, with over 60\% of US students now exhibiting symptoms of mental health conditions \cite{lipson2022trends}. These conditions can often be exacerbated by the stressful and intense nature of higher education. 

One of the most popular and widely supported methods to address mental health challenges is Cognitive Behavioral Therapy (CBT), which focuses on recognizing and addressing unhelpful thoughts and behaviors \cite{beck2020cognitive}. 
CBT often involves completing daily ``homework", such as at-home exercises, as a component of care, since practicing new techniques and applying new ideas through completing homework leads to better patient outcomes \cite{kazantzis2010meta}. However, CBT homework can be hard for therapists to assign and oversee \cite{prasko2022homework} and for clients to complete, due both to internal factors, such as a client’s views towards homework and their motivation to address cognitive distortions, as well as external factors, such as convenience and the amount of time required. 
 Combined with the recent breakthroughs in large language models (LLMs), which have been applied in the mental health domain with some studies reporting positive results \cite{de2023benefits}, an ongoing area of research is exploring how socially assistive robots (SARs) can be leveraged to support CBT \cite{dino2019delivering, kian2024can}.
However, 
further research is needed to ensure safe and effective mental health support.

Motivated by this need for addressing the mental health crisis and the active research on evaluating the use of LLMs in the mental health domain \cite{cho2023evaluating},
in this work we assess the performance of physically embodied and disembodied LLM-powered agents for supporting CBT exercises.
Specifically, we analyze transcripts collected from a two-week study in which 26 university student participants completed daily interactive CBT exercises in their residences using either an LLM-powered physical SAR or a disembodied chatbot. We found that interacting with an embodied agent significantly affected participants’ intimate disclosures of their opinions, judgments, and emotions, with the robot condition having a significantly higher percentage of high-intimacy disclosures than the disembodied chatbot condition. 
Furthermore, both evaluative intimacy and engagement significantly increased over time in the embodied robot condition but decreased in the chatbot condition.
Our findings highlight the role of the SAR's physical embodiment in engaging and facilitating productive conversations with the study participants and supporting their at-home CBT homework exercises.

\section{Background and Related Work}
\subsubsection{Chatbots as Therapeutic Systems}

Chatbots have a long history of therapeutic applications \cite{weizenbaum1966eliza}. A review of chatbots and conversational agents used in therapeutic studies reveals that by providing task-oriented exercises in a dialog format, chatbots can be beneficial for people who struggle to reveal their distress to other people \cite{vaidyam2019chatbots}. They can also be used to support CBT interventions \cite{omarov2023artificial}. Recently, Tessa, a rule-based chatbot designed to help users address eating disorders, demonstrated that while such systems show promise, they are limited in their ability to be spontaneous and flexible \cite{chan2022challenges}. However, with the rise of LLMs capable of human-like dialog generation, there has also been rapid growth in utilizing LLMs in therapeutic chatbots for more flexible conversations \cite{ma2023understanding}. 

LLMs have vastly expanded the capabilities of interactive technologies, including chatbots \cite{yigci2024large} and socially interactive robots \cite{shi2024largelanguagemodelsenable}, across a variety of domains.  For instance, LLM-powered systems can augment traditional therapy as an adjunct to sessions with a therapist \cite{cho2023evaluating} and help mitigate loneliness through conversation \cite{jo2024understanding}. LLMs support many applications in the mental health domain \cite{lawrence2024opportunities} including education, assessment, and interventions. 

\subsubsection{Socially Assistive Robots as Therapeutic Systems}
SARs assist people through social non-physical interactions \cite{feil2005multi} and have been applied to deliver CBT exercises \cite{dino2019delivering, kian2024can}. Research has shown that human users are responsive to robots’ “non-verbal and verbal signals, social language, and conversational gestures” as they capture the user’s attention more often \cite{sidner2004look}. Previous studies have demonstrated that a SAR's physical embodiment leads to increased user engagement and enjoyment \cite{mataric2019human, dennler2021personalizing}. SARs have also been applied to support mental health outcomes by guiding mindful exercises \cite{shi2023evaluating}, delivering motivational interviews \cite{galvao2018experiences}, 
among other uses.

Moreover, embodied agents can elicit intimate self-disclosures from people in therapeutic interactions \cite{soleymani2019multimodal,birmingham2020can}. Expressing feelings to a robot may be even more beneficial to health outcomes compared to a written form of self-reflection \cite{li2023tell}. Embodied robots can facilitate rapport building \cite{deng2019embodiment}, which, in turn, increases self-disclosure \cite{dianiska2021using}. Research has also shown that people may feel more comfortable sharing their intimate occurrences with a robot than a human therapist \cite{khawaja2023your}.  These capabilities make SARs a natural candidate for supporting the therapy process.

\subsubsection {Engagement in Therapy}

Engagement encompasses attention to and investment in the therapeutic process \cite{tetley2011systematic} and is an important predictor of success in therapy, as one must be attentive and invested in the therapeutic process to see results \cite{bolton2003therapeutic}. The patient’s involvement in the therapeutic process and overall tendency for earnest self-disclosure leads to positive therapeutic outcomes \cite{farber2003patient}, whereas low engagement leads to poor therapeutic outcomes \cite{o2009disengagement}. 
SARs have shown promise in cultivating engaged relationships, i.e. they were more robust in task-based interactions than an animated character and around the same level as humans \cite{kidd2004effect}, and a recent review found that SARs increase engagement in a wide variety of therapeutic relationships and settings \cite{riches2022therapeutic}.

 \subsubsection{Intimacy in Therapy} 

 Another measure of self-disclosure in therapy is conversational intimacy, which is operationalized along two dimensions: descriptive and evaluative intimacy \cite{morton1978intimacy}. Descriptive intimacy is the disclosure of facts about oneself, and evaluative intimacy is the disclosure of opinions, judgements, and emotions. Evaluative intimacy tends to be more important than descriptive intimacy in building a strong therapeutic relationship because it focuses on the expression of emotions, values, and personal judgments, all of which are central to fostering trust, empathy, and connection \cite{morton1978intimacy}. Evaluative intimacy plays a critical role in sharing strong emotions and personal judgments and allows clients to feel understood and supported; it involves revealing vulnerabilities crucial for healing and growth \cite{hornstein1988development}. In contrast, descriptive intimacy, while important, primarily focuses on the exchange of factual information without necessarily involving the emotional depth that evaluative intimacy provides. Therefore, while both forms contribute to relationship-building, evaluative intimacy is more central to forming a deep, healing connection \cite{laurenceau2004intimacy}. 

\subsubsection{Therapy Duration}
The long-term efficacy of CBT has been established by countless trials and systematic reviews \cite{chambless2001empirically,norton2007meta}. Duration of therapy significantly affects its outcomes \cite{seligman1995effectiveness}, with long-term exposure suggesting increased benefits. However, while solution-focused therapy can span a multitude of months or years, short-term therapies have been shown to produce benefits more quickly \cite{knekt2008randomized}. The transition from efficacy-focused to effectiveness-focused assessments has motivated many clinical trials that have successfully documented the effectiveness of short-term CBT treatment \cite{dimauro2013long}. Specifically, research has demonstrated that eight CBT sessions are sufficient in producing effective outcomes as well as satisfaction in patients \cite{meyer2022feasibility}. 

While there are some longitudinal studies assessing the effects of chatbot-administered support (e.g., to improve mental well-being \cite{kleinau2024effectiveness}) and the use of SARs for continued support after illness (e.g. short-term rehabilitation \cite{feingold2024socially}), there is a dearth of research assessing both chatbot and SAR outcomes over time with respect to disclosures in CBT exercises. To address this research gap, we devised an experimental setup that examined each technology's impact on engagement and intimacy, respectively, in administering daily CBT exercises over the course of a short-term study. 

Based on the literature reviewed above, we postulate that: a) the embodiment of a SAR outperforms a non-embodied chatbot in terms of the measured outcomes of interest; b) evaluative intimacy assessments respond to therapeutic treatment more than descriptive intimacy; c) the discussed measures of self-disclosure (engagement, intimacy) increase over time through exposure to CBT exercises; and d) the aforementioned effects interact with each other. We thus formulate the following hypotheses:

\noindent
\textbf{H1}: Over time, participants will increase self-disclosures of their thoughts, opinions, and feelings (evaluative intimacy) in the embodied robot condition more than in the chatbot condition.

\noindent
\textbf{H2a}: Over time, participants will increase self-disclosures of personal details about themselves (descriptive intimacy) in the embodied robot condition more than in the chatbot condition.

\noindent
\textbf{H2b}: Over time, CBT exercises will affect evaluative intimacy outcomes more than descriptive intimacy outcomes. 

\noindent
\textbf{H3}: Over time, participants will increase their engagement with the exercise in the embodied robot condition more than in the chatbot condition.

\section{Methodology}
To address the above hypotheses, we analyzed the transcripts of CBT sessions from two experimental conditions conducted over two weeks. Participants’ text responses to LLM-led CBT exercises were prepared according to content analysis methodology and statistically analyzed using 2 x 2 factorial mixed ANOVAs.

\subsection{Study Setup}

\subsubsection{Experiment Design \& Procedure}
The study featured two conditions: participants were assigned to complete CBT exercises with either an LLM-powered embodied SAR (referred to as the robot or embodied condition) or with an LLM-powered disembodied chatbot (referred to as the chatbot or disembodied condition). This study was approved by the university's Institutional Review Board (IRB UP-22-01080). Participants were informed that CBT exercises must be completed daily and were sent daily reminders to complete their exercises throughout the study. Exercise completion was compulsory on days 1-8 of the study and optional on days 9-15. Participants interacted with the system using a web application that was securely hosted on Amazon Web Services (AWS). They logged into the platform with unique credentials and were prompted to give their consent for being recorded prior to each interaction. 
In both conditions, the research team presented a mock session demonstrating how to navigate the platform prior to the start of the study. Participants in the robot condition were provided with a Blossom robot, speaker, and charger that the research team helped to set up in participant homes. All participants were instructed to contact the study coordinator if they had any problems or concerns. The daily CBT sessions were expected to last approximately half an hour. 

\begin{figure}[h]
    \centering
     \subfloat{{\includegraphics[width=7cm]{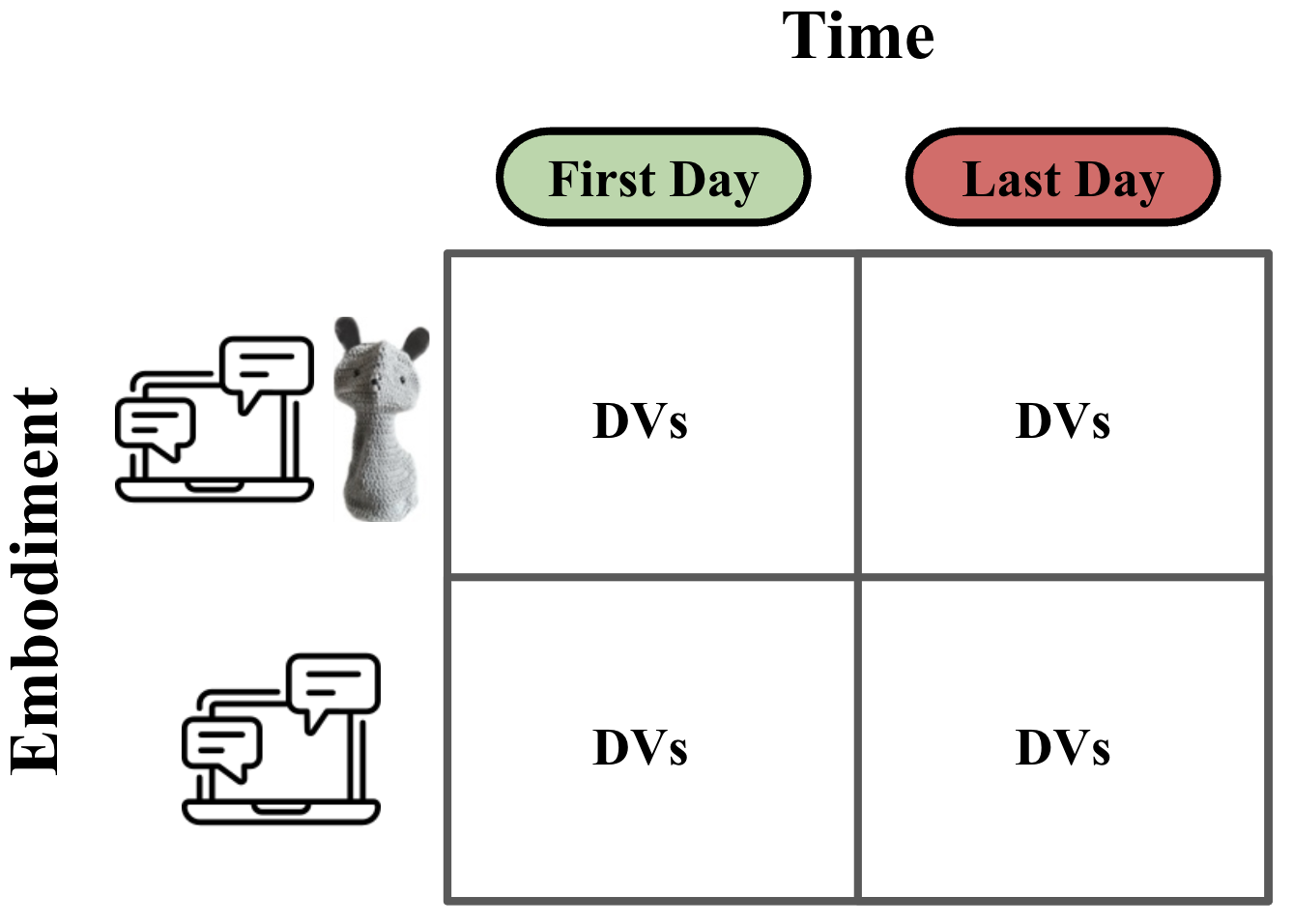} }}%
    \caption{Factorial ANOVA. A two-way mixed design assessing the relationship between embodiment and time on the dependent variables (DVs): evaluative intimacy, descriptive intimacy, and engagement.}%
    \label{fig:pe-linechart-interaction}%
\end{figure}


\subsubsection{Prompt Generation}
A set of interactive exercises based on traditional CBT homework assignments were developed. For each exercise, a series of prompts were crafted to instruct the LLM on how to guide a user through the exercise. 
Through in-lab testing with researchers on the project, a subset of CBT exercises was selected: Cognitive Restructuring \cite{aid_2017}, for identifying cognitive distortions and practicing altering thoughts toward a healthier perspective; and Coping Strategies \cite{ackerman_2017}, for identifying ways to manage stressors to support healthier, more productive behavior. Zero-shot prompting using GPT-3.5 was used for the study.

\subsubsection{Robot Design}
The embodied condition used the Blossom robot \cite{shi2024build, suguitan2019blossom} with a neutral gray crocheted exterior and the ‘Joanna’ voice from AWS Polly.
When the robot was switched on, it transitioned between the speaking state--in which it verbalized the chat messages generated by the LLM--and the idle state--in which the robot waited for the participant to respond. In the speaking state, the robot performed horizontal head movements synchronized with its speech. 
In the idle state, it performed vertical head movements that mimicked a breathing motion similar to breathing in deep-breathing relaxation exercises.

\subsubsection{Participants}
Participants were recruited among the university student population through emails to university departments and an e-bulletin post on a study promotion board. Recruited participants met the following inclusion criteria: over 18 years of age, proficient in English, with normal or corrected-to-normal vision and hearing. The PHQ-9 measure of depression \cite{kroenke1999patient} was used to screen out prospective participants with PHQ-9 scores of 15 or higher, indicating depression. Participants were given initial recruitment forms and were made aware of mental health resources available on campus. They attended a one-on-one informed consent meeting where they chose to sign the consent form. After screening, 28 students were selected to participate in the study. Two participants dropped out shortly after the start of the study, resulting in a total of 26 participants: 14 in the robot condition and 12 in the chatbot condition. Their ages ranged from 18 to 25 years old. Participants self-identified as men (10), women (13), or preferred not to answer (3). Six of the participants were graduate students and 20 were undergraduates. All received a US \$150.00 Amazon gift card after the study was completed. 
	

\subsection{Data Annotation}

We applied content analysis methodology to guide data annotation and analysis \cite{krippendorff04, wimmer1987mass}. 


\subsubsection{Annotation Scheme}
To quantify the participants' responses, we developed an annotation scheme to categorize the data. 
We defined the unit of analysis as each participant response in the dialogue with the LLM, i.e., each turn (which ranged from one to several sentences) that is preceded and/or followed by an utterance from the LLM. For each unit, we coded three variables of interest: the level of conversational intimacy assessed along the two dimensions of descriptive and evaluative intimacy (high or low for each, as in \cite{morton1978intimacy}), and how engaged the participant was with the exercise (active or passive, as in \cite{nguyen2018understanding}). 

\subsubsection{Annotator Training}
Four annotators (undergraduate student researchers) were trained to evaluate the participants' responses. Over a two-week training period, the research team conducted four workshops to train annotators in using the coding scheme and provided check-ins to discuss questions. To test intercoder reliability (ICR), we randomly selected over 10\% of the transcripts, per \cite{neuendorf2017content}. The average ICR for all four annotators was 81.92\% (78.89\% for descriptive intimacy, 79.51\% for evaluative intimacy, and 87.35\% for engagement). The average Cohen's kappa score was substantial \cite{viera2005understanding}: $\kappa$ = 0.603 (0.57 for descriptive intimacy, 0.63 for evaluative intimacy, and 0.61 for engagement).

\subsubsection{Data Preparation}
For each participant, we extracted the following metrics from each session’s annotated transcript: percentage of high evaluative intimacy, percentage of high descriptive intimacy, and percentage of active vs. passive engagement. 



\section{Results}

\subsection{Evaluative Intimacy}

A two-way mixed ANOVA with one between-subjects factor (condition) and one within-subjects factor (time) was devised to assess the main effects of embodiment (robot vs. chatbot) and time (first day vs. last day) as well as their interaction effect on evaluative intimacy. The factorial ANOVA utilized type III sum of squares, which are preferable to type II for unequal group sizes and when an interaction is present \cite{field2012discovering}.

\begin{figure}[h]
    \centering
    \subfloat{{\includegraphics[width=7cm]{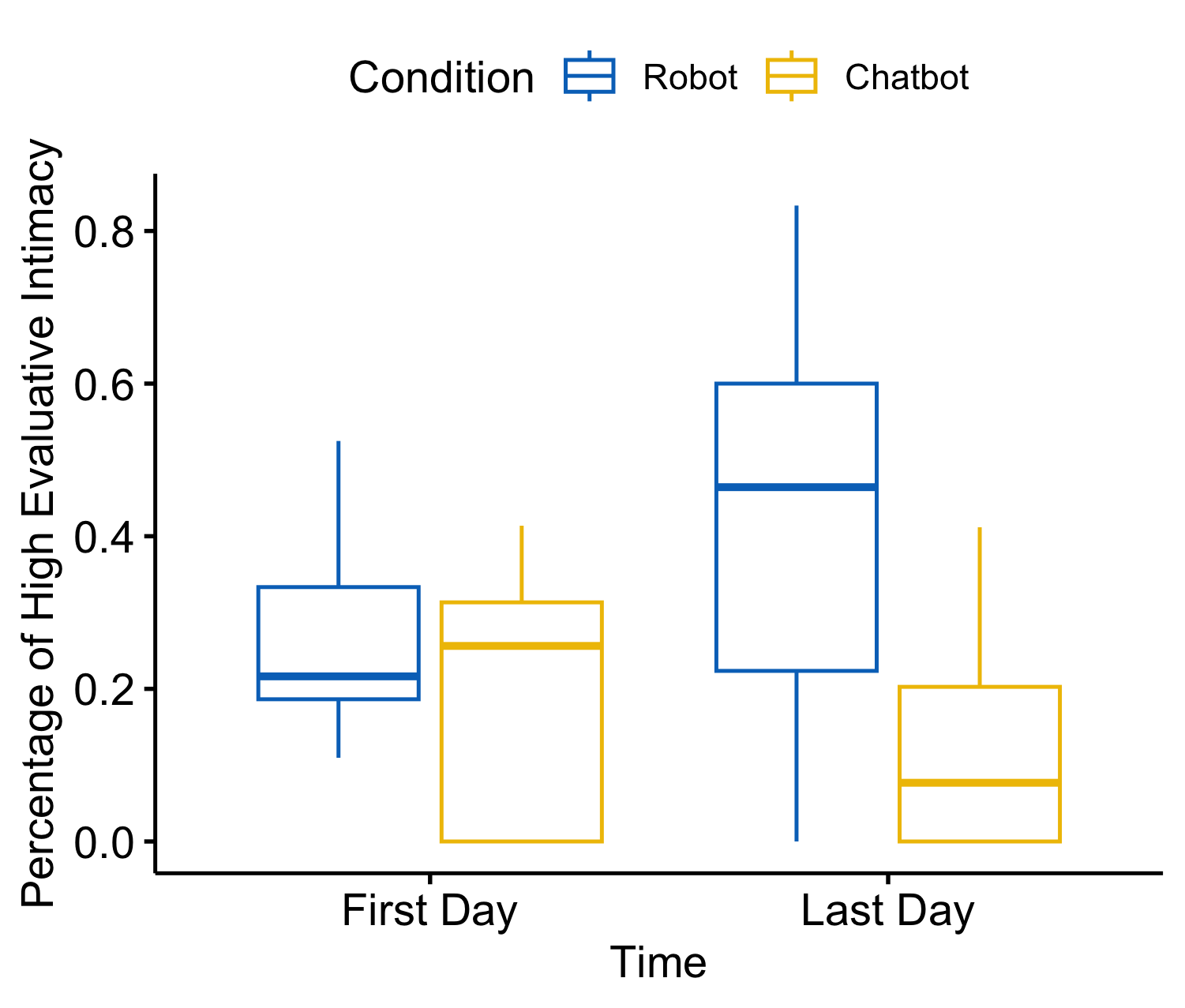} }}
    \qquad
    \subfloat{{\includegraphics[width=7cm]{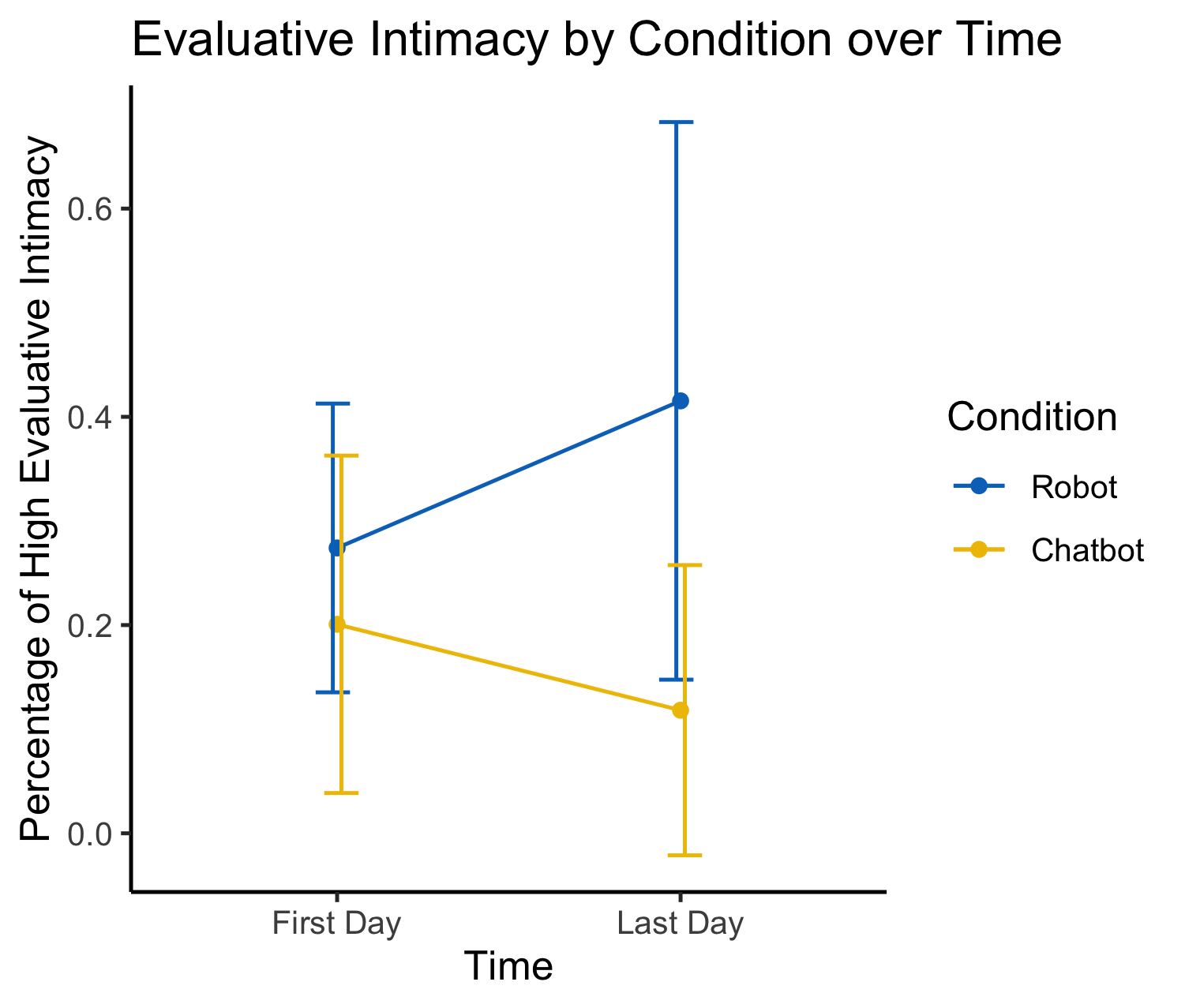} }}%
    \caption{Percentage of High Evaluative Intimacy Disclosures by Condition Over Time (top) Boxplot illustrating the median and interquartile range of the distribution across conditions on the First and Last Days (bottom) Line plot depicting the mean percentage of disclosures over time by condition, with non-parallel trends suggesting a potential interaction effect.}%
    \label{fig:pei-linechart-interaction}%
\end{figure}

The variables were normally distributed after winsorizing; Levene’s test for the between-subjects variable was not significant. The main effect of embodiment was statistically significant at $p < .01$ and yielded an effect size of 0.21 (\emph{F}(1, 24) = 9.23, $p = 0.006$). The mean evaluative intimacy in the robot condition (\emph{M} = 0.35, \emph{SD} = 0.24) was significantly higher than in the chatbot condition (\emph{M} = 0.16, \emph{SD} = 0.15)
. The main effect of time was not significant, however, means indicated a trend towards higher overall evaluative intimacy on the last day of the study (\emph{M} = 0.28, \emph{SD} = 0.26) compared to the first day (\emph{M} = 0.25, \emph{SD} = 0.19)
.

The interaction effect of embodiment by time was significant at $p < .05$ with an effect size of 0.09 (\emph{F}(1, 24) = 7.30, $p = 0.01$). Participants in the robot condition experienced an increase in evaluative intimacy from the first day (\emph{M} = 0.27, \emph{Md} = 0.22, \emph{SD} = 0.14) to the last day (\emph{M} = 0.42 , \emph{Md} = 0.46, \emph{SD} = 0.27), while participants in the chatbot condition experienced a decrease from the beginning (\emph{M} = 0.20 , \emph{Md} = 0.26, \emph{SD} = 0.16) to the end of the study (\emph{M} = 0.12 , \emph{Md} = 0.08, \emph{SD} = 0.14), see Figure \ref{fig:pei-linechart-interaction}.

\subsection{Descriptive Intimacy}

Another 2 x 2 ANOVA was run to predict descriptive intimacy. The assumptions of normality and homogeneity of variance were met. The main effect of embodiment was not significant, yet means indicated that the participants in the robot condition had higher descriptive intimacy (\emph{M}=0.29, \emph{SD}=0.18) than participants in the chatbot condition (\emph{M}=0.21, \emph{SD}=0.21). The main effect of time was not significant. The means indicated that there was a slight drop in descriptive intimacy from first day (\emph{M}=0.26, \emph{SD}=0.17) to last day (\emph{M}=0.25, \emph{SD}=0.22).
Lastly, the interaction effect of embodiment and time was not significant.

\begin{figure}[h]
    \centering
    \subfloat{{\includegraphics[width=6cm]{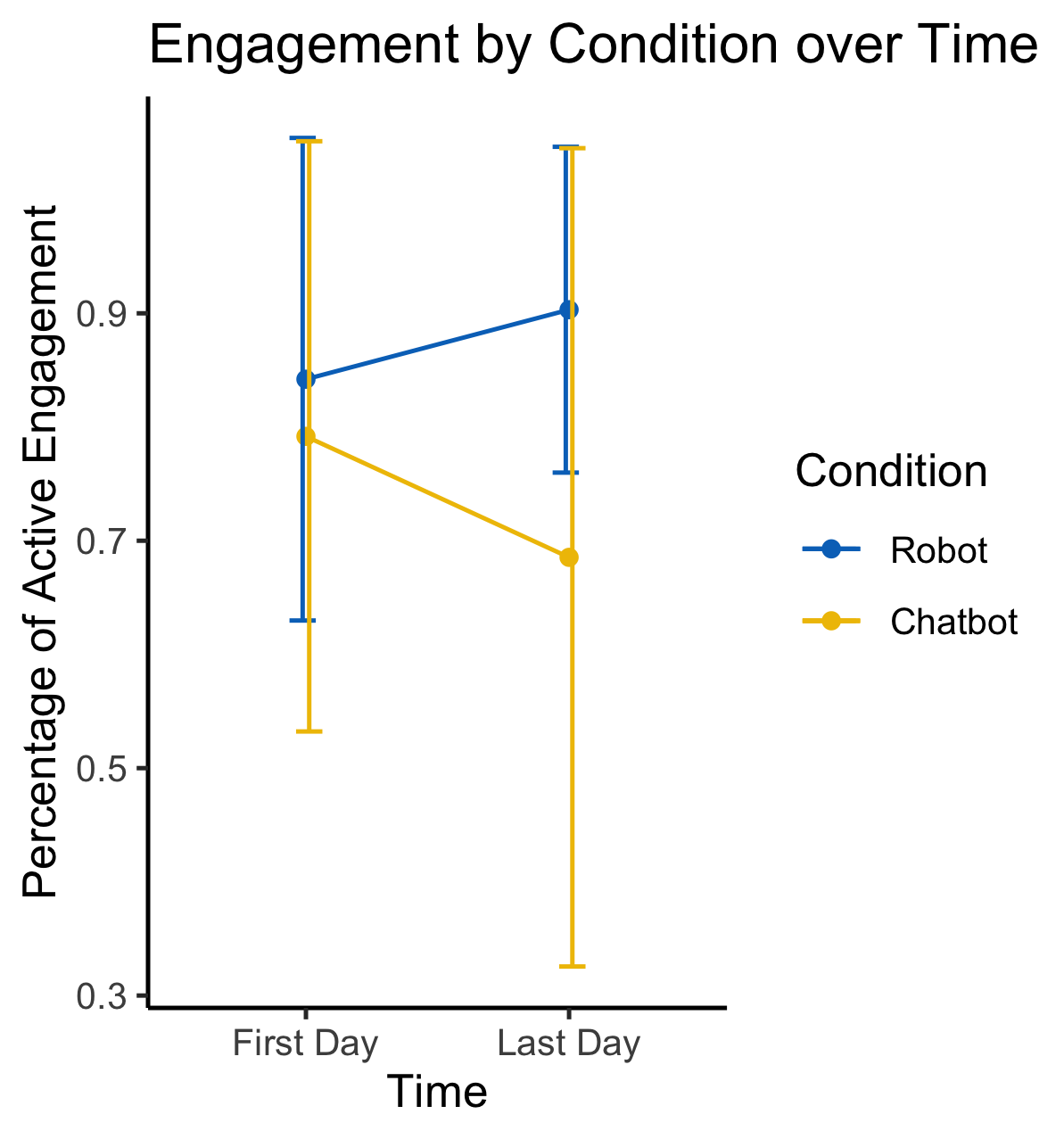} }}%
    \caption{The Percentage of Active Engagement by Condition over Time. Line plot depicting the mean percentage of engaged disclosures over time by condition, with non-parallel trends suggesting a potential interaction effect.}%
    \label{fig:pe-linechart-interaction}%
\end{figure}

\subsection{Engagement}

A third two-way mixed ANOVA was run to assess main effects and interaction effect of embodiment and time with respect to engagement.
Post-winsorization, the variables were normally distributed, and Levene’s test confirmed that they had equal variances. The main effect of embodiment was not significant; however, means indicated that the participants in the robot condition had higher engagement (\emph{M} = 0.87, \emph{SD} = 0.18) than the participants in the chatbot condition (\emph{M} = 0.74, \emph{SD} = 0.31)
. The main effect of time was not significant.

The interaction effect of embodiment and time was significant at $p < .05$ with an effect size of 0.03 (\emph{F}(1, 24) = 5.14, $p = 0.03$). Participants in the robot condition experienced an increase in engagement from the first day (\emph{M} = 0.84, \emph{Md} = 0.93, \emph{SD} = 0.21) to the last day (\emph{M} = 0.9, \emph{Md} = 1, \emph{SD} = 0.14), while participants in the chatbot condition experienced a decrease from the beginning (\emph{M} = 0.79, \emph{Md} = 0.89, \emph{SD} = 0.26) to the last day (\emph{M} = 0.69, \emph{Md} = 0.85, \emph{SD} = 0.36), see Figure \ref{fig:pe-linechart-interaction}.

\section{Discussion}
Time plays an important role in therapy; improvements are rarely immediate and meaningful change does not happen overnight, as many decades of human-led therapy have shown \cite{seligman1995effectiveness}. Our study investigated the effect of time and embodiment on LLM-led therapeutic exercises.
We compared the ability of an embodied SAR and a disembodied chatbot to deliver CBT exercises over the course of two weeks. Our findings show that the SAR elicited both greater participant engagement and greater levels of evaluative intimacy in disclosures than the chatbot.


Prior research has shown that humans feel comfortable sharing intimate experiences with robots \cite{khawaja2023your}; we built on those findings by further delineating disclosure intimacy into two dimensions, descriptive and evaluative, following a framework by \cite{morton1978intimacy} and applying it to a Human-Robot Interaction (HRI) context. Transcripts of participants in the embodied SAR condition revealed that the percentage of high evaluative intimacy statements increased significantly from the beginning to the end of the study, while levels in the disembodied chatbot condition decreased.  Therefore, {\bf H1} is supported. Our finding of the positive impact of embodiment adds to a large body of work that has shown that embodied robots elicit smoother interactions, have access to different social cues, and provide a greater sense of social presence than non-embodied interfaces \cite{cassell_bickmore_hannes_högni_vilhjálmsson_yan_2000, wallkotter_tulli_castellano_paiva_chétouani_2021, lee_jung_kim_kim_2006} and elicit increased intimacy over time \cite{farber2003patient}. 

In contrast, there was a negative development of intimate disclosures in the chatbot condition, consistent with study results reporting loss of interest in interactions with chatbots in longitudinal contexts \cite{folstad2020users}. There is evidence that social processes shown to aid relationship formation decrease after multiple interactions with a chatbot \cite{longitudinal2021}, which may contribute to these outcomes. 

While evaluative intimacy involves expressions of emotional truths, judgments, and vulnerabilities, which are crucial when developing trust and connection with others \cite{hornstein1988development}, descriptive intimacy focuses on describing facts. Although our results showed that system embodiment significantly interacted with time when predicting evaluative intimacy, embodiment had no significant effect on descriptive intimacy. Therefore, {\bf H2a} was not supported. Data pertaining to descriptive intimacy showed similar trends to the assessment of evaluative intimacy, but were weaker in effect and not significant, confirming our assumptions about the heightened role of evaluative conversational disclosures over descriptive ones based on prior research \cite{laurenceau1998intimacy,duck1988handbook,sullivan1953interpersonal}. Therefore, {\bf H2b} was supported.

The embodiment of the LLM also interacted significantly with the study duration relative to participant engagement. Over time, the percentage of actively engaged responses increased in the robot condition and decreased in the chatbot condition. Therefore, {\bf H3} was supported. A review of engagement in longitudinal SAR studies \cite{oertel2020engagement} determined that, for long-term interactions to be successful, the robot must be emotionally aware and generate responses adapted to the conversational context \cite{ahmad2017adaptive}. Following those criteria for successful engagement, we were able to replicate those findings in our study. However, the chatbot  saw a drop in the participants' engagement. SARs have been found to outperform other embodiments in terms of engagement; for instance, they compared favorable to a virtually embodied agent when measuring participants involvement in human-agent interactions \cite{kidd2004effect}. Our findings confirm the favorable role of embodied SARs in comparison with disembodied chatbots with respect to engagement.


Our results suggest that, in longitudinal studies, embodiment is critical for encouraging engagement and evaluative intimacy to support mental health outcomes. These findings are especially relevant to the mental health domain as most mental health issues require continued support \cite{kazantzis2010meta}. The outcomes of this work aim to advance access and scalability of mental health services through the use of assistive technologies. Specifically, we suggest that LLM-powered SARs can be an important component of delivering mental health interventions and complementing the work of human therapists as supplementary tools, by supporting at-home practice of CBT and other mental health interventions, all of which benefit from long-term training and practice to be effective.
Future studies should extend these findings to other populations to support generalization beyond the university student context and examine varying study durations to contribute to the shared understanding of the supportive role of SARs in the mental health domain.

\section{Conclusion}
To understand the effect of embodiment on LLM-powered agents in the context of supporting CBT practice, we analyzed transcripts collected during a 15-day study in which university students completed CBT exercises with either an LLM-powered chatbot or SAR. We discovered that between the participants' first and last days of interacting with the agent, the percentage of actively engaged responses and their evaluative intimacy increased in the SAR condition and decreased in the chatbot condition. These results demonstrate that for long-term at-home support for CBT exercises, an embodied agent is likely to improve the performance of key therapeutic measures, while a chatbot alone may not suffice. Therefore, developers of therapeutic technologies for mental health should keep in mind the benefits of embodiment as they design systems for long-term support.

\bibliographystyle{IEEEtran}
\bibliography{base}

\end{document}